\documentclass[10pt,twocolumn,letterpaper]{article}
\pdftrailerid{}

\usepackage[accsupp]{axessibility}
\usepackage{amsmath}
\usepackage{amssymb}
\usepackage{graphicx}
\usepackage{xcolor}
\usepackage{algorithm}
\usepackage{algpseudocode}

\usepackage{wacv}

\definecolor{wacvblue}{rgb}{0.21,0.49,0.74}
\usepackage[pagebackref,breaklinks,colorlinks,allcolors=wacvblue]{hyperref}

\def\wacvPaperID{563}
\def\confName{WACV}
\def\confYear{2027}

\title{Single-Annotation, Multi-Hypothesis Segmentation with Semantic-Spatial Residual Banks}

\author{Armin Dadras$^{1,2}$ \quad Robert Sablatnig$^1$ \quad Franziska Proksa$^2$ \quad Markus Seidl$^2$\\
$^1$Computer Vision Lab, TU Wien, Vienna, Austria\\
$^2$UAS St. P\"olten, St. P\"olten, Austria\\
{\tt\small adadras@cvl.tuwien.ac.at}}

\hypersetup{pdftitle={Single-Annotation, Multi-Hypothesis Segmentation with Semantic-Spatial Residual Banks},pdfauthor={Armin Dadras; Robert Sablatnig; Franziska Proksa; Markus Seidl},pdfsubject={WACV 2027 Algorithms Track},pdfkeywords={single-annotation segmentation, multi-hypothesis segmentation, semantic-spatial residual bank}}
\begin{document}
\maketitle

\begin{abstract}
Multi-rater segmentation datasets show that a single image can admit several
valid masks, yet most training sets expose only one annotation per image. We
study single-annotation multi-hypothesis segmentation: each training image
exposes one fixed mask, while the model must return a finite bank that covers
held-out human annotations at test time. Our experiments reveal a
representation tradeoff. Banks learned directly from images recover variations
in local boundaries and extent but lose structural hypotheses, whereas frozen
semantic features preserve structural grouping but underrepresent spatial
variation. We introduce the Semantic-Spatial Residual Bank (SSRB), which
freezes a semantic bank trained with balanced assignments and learns a
residual from the image for each hypothesis. Across five medical endpoints, SSRB
improves coverage of held-out masks at $K=8$ over a Probabilistic U-Net under
the same one-mask constraint. On LIDC and LIDC-hard, it also produces
substantially more distinct useful masks. On BSDS, it preserves the semantic
bank's performance in grouping regions. Its advantage grows as more masks
are returned, showing that residuals learned for corresponding heads improve
coverage of local annotation variation while preserving how regions are grouped.
\end{abstract}

\section{Introduction}
\label{sec:intro}

A single image can admit several defensible segmentations. Radiologists may
draw different extents of one lung nodule, ophthalmologists may disagree on an
optic-cup boundary, and people may parse a natural image into different
regions~\cite{armato2011lidc,almazroa2018riga,abhishek2025imapp,arbelaez2011bsds}.
Collecting several masks for every image is costly, however, and many training
sets expose only one. A useful model should still return plausible alternatives
for a new image without having seen several labels for any training example.

We study this problem with one fixed annotation per training image. Given a
new image, the model returns $K=8$ masks to cover the different segmentations
that annotators might produce.

A finite bank can be sampled from a probabilistic decoder or emitted directly
by several heads~\cite{kohl2018probabilistic,baumgartner2019phiseg,
guzman2012mcl,lee2016smcl}. Direct prediction is simple, but one representation
does not serve every kind of ambiguity equally well. Raw image features retain
fine boundaries and local extent changes. Frozen semantic features retain
object and region structure. With one annotation per image, our two controlled
banks expose the conflict: the raw bank is much stronger on fine medical
segmentations, whereas the semantic bank is much stronger on the task of grouping regions more broadly.

\begin{figure*}[t]
\centering
\includegraphics[width=.95\textwidth]{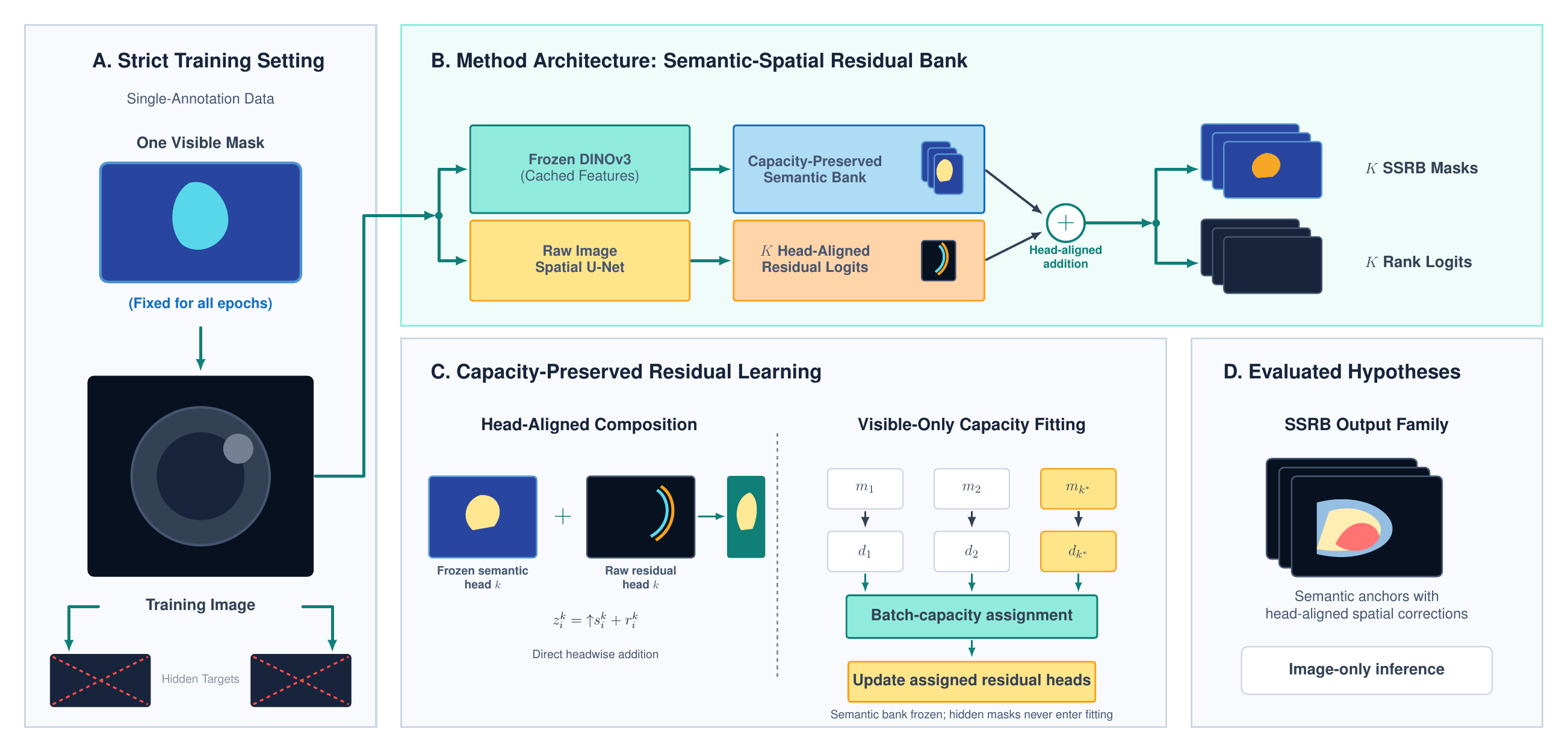}
\caption{\textbf{Overview of the Semantic-Spatial Residual Bank.} One fixed
mask is visible during fitting. A
bank trained with balanced assignments over frozen DINOv3 features supplies
$K$ semantic anchors. A U-Net takes the image and learns one spatial residual for each head,
and adding their logits produces the final bank.}
\label{fig:ssrb}
\end{figure*}

We therefore introduce the \emph{Semantic-Spatial Residual Bank} (SSRB; Fig.~\ref{fig:ssrb}). We
first train an eight-head bank over frozen DINOv3
features~\cite{simeoni2025dinov3}, assigning images to heads with balanced capacities through a linear
assignment~\cite{kuhn1955hungarian}. We then freeze that bank and train
a small raw-image U-Net~\cite{ronneberger2015unet} to correct each head's mask
and rank logits. Each semantic hypothesis anchors one spatial correction. Adding their logits
produces eight masks, using the same rule on every endpoint.
Across IMA++~\cite{abhishek2025imapp}, LIDC~\cite{armato2011lidc}, and
separate RIGA disc and cup tasks~\cite{almazroa2018riga}, SSRB improves
$K=8$ coverage over a Probabilistic U-Net~\cite{kohl2018probabilistic} trained with the same visible masks,
with simultaneous confidence bounds above zero. It matches the raw bank on
LIDC and the semantic bank on BSDS~\cite{arbelaez2011bsds}, while improving
both branches on IMA++ and RIGA. The pattern also holds with
DINOv2~\cite{oquab2023dinov2}. Across bank
sizes, the improvement emerges as additional useful hypotheses are admitted.

Our contributions are:
\begin{itemize}
\item We study single-annotation multi-hypothesis segmentation,
training on one mask per image and evaluating against held-out alternatives.
\item We identify a representation tradeoff: end-to-end spatial banks recover
local extent variation but lose structural groupings, while frozen semantic
banks preserve structure but miss local variation.
\item We introduce SSRB, which freezes a semantic bank trained with balanced
assignments and learns a residual from the image for each head. Every endpoint uses the same
architecture and composition rule.
\item On IMA++, LIDC, RIGA, and a BSDS-derived task, we show that SSRB retains
the stronger representation and improves coverage of medical annotations over a
Probabilistic U-Net trained under the same one-mask constraint.
\end{itemize}

\section{Related Work}
\label{sec:related}

\paragraph{Ambiguous segmentation.}
Probabilistic U-Net~\cite{kohl2018probabilistic}, PHiSeg
~\cite{baumgartner2019phiseg}, stochastic segmentation networks
~\cite{monteiro2020stochastic}, and diffusion models
~\cite{rahman2023ambiguous,christensen2025diffusionbased} represent ambiguous
predictions through latent variables, distributions, or repeated samples.
These methods are commonly evaluated with, and can exploit, repeated
annotations. We compare their banks of predictions when training provides one fixed mask
per image, including samples from the Probabilistic U-Net prior and the
released CIMD model.
Since
recorded disagreement can also contain noise~\cite{kimhi2025noisy}, we report
coverage together with unsupported masks and redundancy.

\paragraph{Direct multi-hypothesis prediction.}
Multiple-choice learning assigns each target to its closest predictor
~\cite{guzman2012mcl,lee2016smcl}; Deterministic Mode Proposals also predict a
bounded set without iterative sampling~\cite{gerard2026dmp}. When heads compete,
some may receive little supervision. We use an exact capacitated assignment so
that every head trains. This separates head survival from the representation
question addressed by SSRB.

\paragraph{Promptable and query segmentation.}
DETR-style segmenters learn mask queries~\cite{carion2020detr,
cheng2022mask2former}. SAM and SAM~2 produce masks from spatial prompts or
automatic proposal grids~\cite{kirillov2023sam,ravi2025sam2}, while SAM~3 adds
text and exemplar concepts~\cite{carion2026sam3}. EfficientSAM3 distills the
visual and temporal components across the SAM family, and SAM3-LiteText
compresses the text encoder~\cite{zeng2025efficientsam3,zeng2026sam3litetext}.
Ambiguity-aware extensions also include $\mathcal{A}$-SAM~\cite{li2024asam},
which models prompt and object-granularity ambiguity, and
P$^2$SAM~\cite{huang2024p2sam}, which learns probabilistic prompts from
ambiguously annotated medical data. These models use prompts or additional annotations to resolve a requested
object or concept. SSRB predicts eight masks for its trained task directly
from the image. We compare it with automatic SAM~2; the supplement reports
experiments with point prompts separately.

\paragraph{Semantic and spatial features.}
Self-supervised DINO features encode broad semantic structure
~\cite{oquab2023dinov2,simeoni2025dinov3}, whereas convolutional decoders retain
local detail. Multi-path refinement, full-resolution streams, and residual
adaptation of frozen features are established ideas
~\cite{lin2017refinenet,pohlen2017frrn,khan2026keepitfrozen}. SSRB applies
these ideas after the semantic bank has formed its hypotheses. SSRB adds a learned spatial residual to the output logits of each frozen
semantic head, so the two branches refer to the same hypothesis. Training uses
one fixed annotation per image.

\section{Problem Setup and Method}
\label{sec:method}

\subsection{Single-annotation setting}
\label{sec:strict_setting}

For an image (or image--task pair) $x_i$, let
$Y_i=\{y_i^1,\ldots,y_i^{M_i}\}$ denote its human masks. One mask
$y_i^v\in Y_i$ is fixed as visible, and
$Y_i^h=Y_i\setminus\{y_i^v\}$ remains hidden during fitting. Training receives
$(x_i,y_i^v)$; checkpoint selection receives only visible validation masks;
and test inference receives $x_i$ alone. The output is
\begin{equation}
B_i=\{(m_i^k,q_i^k)\}_{k=1}^{K}, \qquad K=8,
\label{eq:bank}
\end{equation}
where $m_i^k$ is a binary mask and $q_i^k$ is a ranking score.

The evaluator reads alternative masks after predictions have been saved and
maps every prediction to the original annotation grid. For a learned logit map $z_i^k$,
\begin{equation}
m_i^k=\mathbf 1\!\left[
\operatorname{BilUp}\!\left(\sigma(z_i^k)\right)\geq .5\right],
\label{eq:evaluation_mask}
\end{equation}
using bilinear interpolation with \texttt{align\_corners=False}. Already
binary external proposals use nearest-neighbor interpolation. RIGA disc and
cup are treated as separate tasks and models.

\subsection{Training the semantic bank}
\label{sec:capacity_anchor}

The semantic bank uses frozen DINOv3 ViT-B/16 patch tokens from zero-based
blocks 5, 8, and 11 at a $28\times28$ grid~\cite{simeoni2025dinov3}. A learned
mixture of the three layers feeds a shared token MLP of width 256 and $K$ mask
heads; pooled tokens feed $K$ rank heads. Dropout is disabled. The trainable bank
has 0.467M parameters. Images are resized to $448^2$. For LIDC, intensities are
clipped to $[-1000,500]$ HU before the stored nodule crop is mapped to three
channels.

For visible mask $y_i^v$, the cost of head $k$ is
\begin{equation}
d_{ik}=\mathcal L_{\rm BCE}(s_i^k,y_i^v)
       +\mathcal L_{\rm Dice}(\sigma(s_i^k),y_i^v),
\label{eq:visible_cost}
\end{equation}
where BCE is the pixel mean and soft Dice uses additive smoothing one.
Ordinary WTA (Winner takes all) can repeatedly choose the same few heads. For a
batch of size $B$, we instead solve
\begin{equation}
\begin{aligned}
\min_{a\in\{0,1\}^{B\times K}}\;&\sum_{i,k}a_{ik}d_{ik},
&\sum_k a_{ik}&=1,\\[-2pt]
&&\sum_i a_{ik}&\in
\{\lfloor B/K\rfloor,\lceil B/K\rceil\}.
\end{aligned}
\label{eq:capacity_assignment}
\end{equation}
A Hungarian solution finds the assignment with the smallest total cost. When $B$ is not
divisible by $K$, the heads receiving one extra item rotate across batches.
The semantic-bank objective is
\begin{equation}
\mathcal L_{\rm sem}=
\frac1B\sum_{i,k}a_{ik}d_{ik}
+\frac{.30}{BK}\sum_{i,k}
\operatorname{BCEWithLogits}(q_i^k,a_{ik}).
\label{eq:capacity_loss}
\end{equation}
The detached assignment supplies a one-hot rank target, so the objective
contains only the assigned segmentation and ranking terms. At validation,
each image independently chooses the head with the lowest cost. Checkpoint
selection uses the objective computed from these choices and the visible masks.

\subsection{Head-aligned spatial residuals}
\label{sec:ssrb}

The $28^2$ semantic field supplies context but may lose local boundary detail.
Let $s_i^k$ and $q_i^{k,S}$ be the mask and rank logits of the trained semantic
bank. A U-Net with four levels and base width 24 takes the image as input and
predicts a mask residual $r_i^k$ and rank residual $\Delta q_i^k$ for each
semantic head. SSRB forms
\begin{equation}
z_i^k=\operatorname{BilUp}_{128}(s_i^k)+r_i^k,
\qquad q_i^k=q_i^{k,S}+\Delta q_i^k.
\label{eq:ssrb}
\end{equation}
Head $k$ in the raw branch therefore corrects head $k$ in the semantic bank.
The anchor is frozen throughout residual fitting. Only the final mask and rank
layers of the raw branch start at zero, so the initial logits equal the
upsampled anchor logits; Sec.~\ref{sec:controls} tests standard output
initialization as well. We evaluate both branches for every image and add their logits with unit
weight.

Residual training recomputes Eq.~\ref{eq:visible_cost} from the fused logits
and uses the same assignment and rank loss as
Eqs.~\ref{eq:capacity_assignment}--\ref{eq:capacity_loss}. Only the 1.094M raw
parameters are updated. The semantic bank uses AdamW with learning rate
$3\!\times\!10^{-4}$, weight decay $10^{-4}$, batch size 32, at most 40
epochs, and patience 8. The residual uses Adam with learning rate $10^{-3}$,
batch size 12, at most 80 epochs, and patience 12.
Both stages select the lowest eligible visible validation objective, breaking
ties in favor of the earlier epoch. Semantic checkpoints become eligible after
four epochs; residual checkpoints are eligible from the first epoch.

Inference runs each branch once, adds logits for corresponding heads, and
returns eight masks in descending score order. Including the frozen DINO
backbone, SSRB has 87.23M parameters, of which 1.56M are task-specific.

\subsection{Evaluation measures}
\label{sec:metrics}

For the first $r$ masks in score order, coverage of the held-out annotations is
\begin{equation}
\operatorname{MLQ}_r(i)=\frac1{|Y_i^h|}\sum_{y\in Y_i^h}
\max_{k\le r}\operatorname{Dice}(m_i^k,y).
\label{eq:mlq}
\end{equation}
MLQ@8 measures oracle coverage: for each reference, it finds the closest mask
in the bank. Worst@8 takes the minimum, rather than the
mean, over hidden masks. Prefix AUC averages MLQ@1 through MLQ@8 and evaluates
ordering. Waste@8 is the fraction of returned masks that have Dice below .30
with every hidden annotation. PairDice and Unique95, the number of clusters after merging
masks with Dice at least .95, describe redundancy. Distinct masks still require assessment of their validity.

\section{Experiments}
\label{sec:experiments}

We compare raw and semantic banks across six endpoints and test whether SSRB
retains their complementary strengths. We evaluate coverage of the bank
against probabilistic, diffusion, and automatic proposal baselines, then
examine direct addition, bank union, initialization, raw-model width, residual
behavior, how coverage changes as more masks are returned, and computation.

\subsection{Data and protocol}
\label{sec:protocol}

We use IMA++ skin lesions~\cite{abhishek2025imapp}, LIDC-IDRI lung nodules
~\cite{armato2011lidc}, and RIGA fundus images~\cite{almazroa2018riga}.
Train/validation/test sizes are 1675/%
240/479 for IMA++,
1378/233/%
267 for LIDC, and
471/123/%
150 for each RIGA task. For LIDC, we segment nodules in crops from their central slices, with patients
separated across splits. Each crop is defined offline from the union of
annotations and supplied to every method. It provides the nodule's location
but does not serve as supervision for alternative masks. RIGA disc and cup use separate
models on shared fundus images. LIDC-hard contains
160 observations from
74 patients.
Which observations are included depends on the seed: the chosen visible mask
must have Dice below .80 with an alternative mask.

\begin{figure}[t]
\centering
\includegraphics[width=\linewidth]{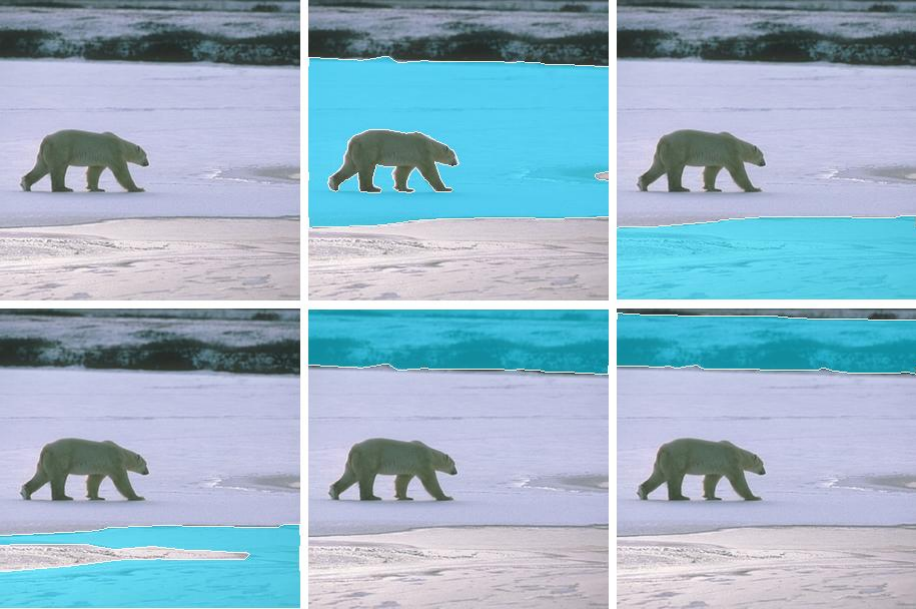}
\caption{\textbf{Building the BSDS masks.} Each image provides a set of binary masks
for connected regions in human partitions. We filter regions by area, merge
near-duplicates, and cap the number of masks.}
\label{fig:bsds_construction}
\end{figure}

BSDS500~\cite{arbelaez2011bsds} supplies a different form of variation. We
turn each human partition into connected-region masks, remove regions below
1.5\% or above 85\% of the image, merge masks at Dice .95, and retain at most
16. The resulting 199/%
100/%
200 split tests how the model groups regions in an image
(Fig.~\ref{fig:bsds_construction}).

All methods are scored through Eq.~\ref{eq:evaluation_mask}. The supplement
provides implementation and split details.

\subsection{Comparisons and inference}
\label{sec:baselines}

The representation study compares a bank over DINOv3 features, a U-Net
that learns a bank directly from images, and SSRB. All three use the same $K=8$ capacity assignment.
We also test a wider raw U-Net with base width 29 instead of 24. Other
comparisons include a single deterministic predictor, MC-dropout
~\cite{gal2016dropout}, ordinary WTA,
and the Rank-Neighborhood objective. All trained baselines receive the same visible annotations as SSRB.
Probabilistic U-Net~\cite{kohl2018probabilistic} draws eight samples from its
prior conditioned on the image. CIMD~\cite{rahman2023ambiguous} produces eight
samples with independently drawn noise, each using 1000 diffusion steps. CIMD takes four input channels. We repeat a grayscale image across these
channels, which may disadvantage it on RGB benchmarks. SAM~2 AMG~\cite{ravi2025sam2} generates proposals from an
internal grid of points; we evaluate up to eight. The supplement also reports
an experiment in which a trained model supplies point prompts to SAM~2.

For SSRB and both representation controls, we use three seeds to choose the
visible masks and three seeds for optimization. We average available metrics
over the choices of visible masks for each source and optimization seed, then
jointly resample optimization seeds and sources.
The source is a patient for LIDC, a raw fundus image for both RIGA tasks, and an
image otherwise. Disc and cup reuse the same fundus draw in each bootstrap
replicate. Paired intervals use 10,000 bootstrap samples. The SSRB versus
Probabilistic-U-Net families use simultaneous 95\% intervals across the five
medical endpoints. Samples from the same probabilistic model are grouped together.

\subsection{Complementary representations}
\label{sec:representation_results}

\begin{table*}[t]
\centering
\small
\setlength{\tabcolsep}{5.0pt}
\begin{tabular}{lcccc}
\hline
Endpoint & DINO Capacity-WTA & Raw Capacity-WTA & SSRB & SSRB Waste@8 \\
\hline
IMA++ & 0.8972
& 0.8994
& \textbf{0.9101}
& 0.0705 \\
LIDC & 0.8325
& 0.8895
& \textbf{0.8932}
& 0.0427 \\
LIDC-hard & 0.8016
& 0.8593
& \textbf{0.8614}
& 0.0452 \\
RIGA-disc & 0.9526
& 0.9576
& \textbf{0.9676}
& 0.0000 \\
RIGA-cup & 0.8886
& 0.9012
& \textbf{0.9141}
& 0.0000 \\
BSDS & 0.5408
& 0.4211
& \textbf{0.5414}
& 0.1726 \\
\hline
\end{tabular}
\caption{\textbf{MLQ@8 across representations.} The raw and semantic banks
have complementary strengths. SSRB retains the stronger branch on LIDC,
LIDC-hard, and BSDS, and improves both branches on IMA++ and the two RIGA
tasks. SSRB values are bold; the last column reports the
fraction of unsupported SSRB masks.}
\label{tab:representation}
\end{table*}

\begin{figure*}[t]
\centering
\includegraphics[width=.94\textwidth]{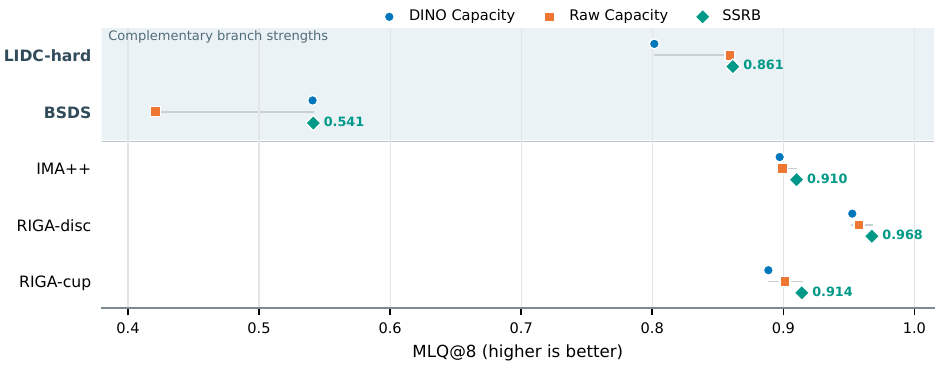}
\caption{\textbf{The representation tradeoff.} Raw features are strongest for
local medical extent, while semantic features are strongest for BSDS grouping.
Learning a residual for each semantic head retains the stronger branch. Values are MLQ@8; BSDS is a
distinct stress test.}
\label{fig:representation_tradeoff}
\end{figure*}

Table~\ref{tab:representation} and Fig.~\ref{fig:representation_tradeoff} show
that the stronger representation depends on the task. The raw bank better
covers variation in nodule extent on LIDC and LIDC-hard, while the DINO bank
better covers the groupings of regions in BSDS. SSRB stays close to the
stronger branch on these three endpoints and improves on both branches on
IMA++ and the two RIGA tasks. Paired intervals for differences from the
stronger branch include zero on LIDC, LIDC-hard, and BSDS (supplement).
The same construction thus supports local variation and broader groupings.

\subsection{Finite-bank coverage}
\label{sec:coverage_results}

\begin{table*}[t]
\centering
\scriptsize
\setlength{\tabcolsep}{2.8pt}
\begin{tabular}{lllccccccc}
\hline
Method & Test input & Bank construction & IMA++ & LIDC & Hard & Disc & Cup & BSDS & ms/image \\
\hline
Probabilistic U-Net & image & 8 prior samples
& 0.8778 & 0.8603
& 0.8136 & 0.9592
& 0.8953 & 0.0600
& 11.57 \\
CIMD & image & 8 diffusion samples
& 0.8496 & 0.5808
& 0.5220 & 0.8098
& 0.8309 & 0.3196
& 57378.97 \\
SAM~2 AMG & image; AMG grid & top $\leq8$ proposals
& 0.3287 & 0.3669
& 0.3452 & 0.1365
& 0.1053 & 0.3044
& 1549.37 \\
\textbf{SSRB} & image & fixed 8 heads
& \textbf{0.9101}
& \textbf{0.8932}
& \textbf{0.8614}
& \textbf{0.9676}
& \textbf{0.9141}
& \textbf{0.5414}
& 45.24 \\
\hline
\end{tabular}
\caption{\textbf{Modern image-only comparisons.} Entries are MLQ at the
reported bank budget. SAM~2 receives no user prompt, but AMG uses an internal
point grid and has variable effective $K$; CIMD uses 1000 diffusion steps per sample. Runtime is end-to-end on one A40 at batch size one.}
\label{tab:modern_baselines}
\end{table*}

SSRB covers the held-out annotations better than Probabilistic U-Net on all five medical
endpoints (Table~\ref{tab:modern_baselines});
the simultaneous confidence intervals for these differences exclude
zero. On LIDC and LIDC-hard, it also returns more distinct masks on average.
The bank thus contains more distinct alternatives with better coverage of the
recorded annotations. The supplement gives paired intervals.

\begin{figure*}[t]
\centering
\includegraphics[width=\textwidth]{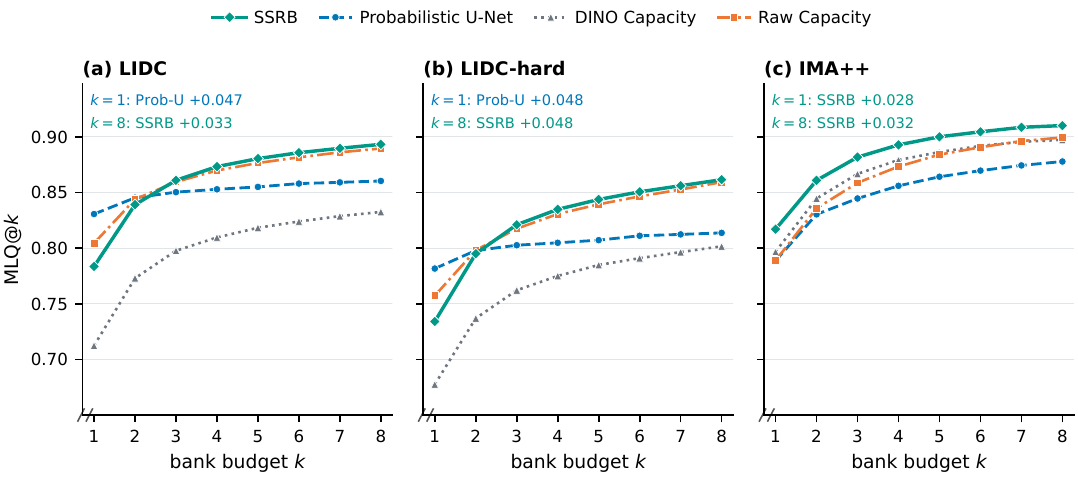}
\caption{\textbf{Coverage as the bank grows.} Absolute MLQ for SSRB,
Probabilistic U-Net, DINO Capacity-WTA, and Raw Capacity-WTA. Probabilistic
U-Net has the stronger first mask on LIDC and LIDC-hard; SSRB crosses it at
$K=3$ and gains as additional hypotheses are admitted. SSRB leads from $K=1$
on IMA++.}
\label{fig:budget_curves}
\end{figure*}

Figure~\ref{fig:budget_curves} shows where the benefit appears as more masks
are returned. Probabilistic U-Net provides the stronger first mask on LIDC
and LIDC-hard, but SSRB overtakes it from $K=3$ onward. On IMA++, SSRB leads
throughout the curve. Broader coverage comes with higher Waste on
IMA++, LIDC, and LIDC-hard: more returned masks have little overlap with any
held-out annotation. Neither method returns such masks on the two RIGA tasks
at the chosen threshold. Probabilistic U-Net remains faster
(Table~\ref{tab:modern_baselines}).

\subsection{Composition controls}
\label{sec:controls}

\begin{table*}[t]
\centering
\scriptsize
\setlength{\tabcolsep}{4.0pt}
\begin{tabular}{lcccc}
\hline
& \multicolumn{2}{c}{LIDC-hard} & \multicolumn{2}{c}{BSDS} \\
Comparison & Control MLQ & SSRB $-$ control [95\% CI]
& Control MLQ & SSRB $-$ control [95\% CI] \\
\hline
Addition after matching heads
& 0.8340
& +0.0274
[+0.0115,
+0.0434]
& 0.3581
& +0.1833
[+0.1715,
+0.1952] \\
Union: four raw $+$ four DINO masks
& 0.8592
& +0.0023
[-0.0063,
+0.0109]
& 0.4584
& +0.0830
[+0.0731,
+0.0928] \\
Tuning the final DINO block
& 0.8018
& +0.0596
[+0.0497,
+0.0695]
& 0.5420
& -0.0006
[-0.0052,
+0.0039] \\
Wider raw U-Net
& 0.8635
& -0.0020
[-0.0130,
+0.0089]
& 0.4203
& +0.1212
[+0.1040,
+0.1383] \\
\hline
\end{tabular}
\caption{\textbf{Composition controls at MLQ@8.} Positive differences favor
SSRB. Learned residuals exceed addition after matching heads. The fixed
$4+4$ union is a descriptive test of bank coexistence. Adapting the final DINO
block does not recover the LIDC-hard gap. The wider raw model approaches SSRB
on LIDC-hard but trails on BSDS. Intervals are simultaneous over each
comparison's reported endpoint family; the supplement reports the full results.}
\label{tab:composition_controls}
\end{table*}

We compare SSRB with addition after matching heads and with a union of masks
from the two banks (Table~\ref{tab:composition_controls}). For addition, we
sum the mask logits of each raw/DINO head pair and average its BCE plus soft
Dice loss over the visible validation set. A Hungarian
assignment~\cite{kuhn1955hungarian} matches the eight heads using these costs.
We then add the mask and rank logits of matched heads. Learning residuals
improves on this control on LIDC-hard and BSDS: training the corrections
helps beyond adding independently trained predictions.

The union sorts each bank by its own rank scores and returns four raw masks
followed by four DINO masks. Scores are not calibrated between banks, and the
allocation is fixed. It reaches a similar mean on LIDC-hard, with an interval
for the difference that includes zero, but falls below SSRB on BSDS. Keeping
half of each bank tests a fixed allocation of the budget between representations.

Conventional initialization of the residual output layers also yields
effective corrections (supplement), so starting these layers at zero is not
essential to the observed gains.

The wider raw U-Net approaches SSRB on LIDC-hard but remains substantially
below it on BSDS, despite having slightly more task-specific parameters.
The paired interval on LIDC-hard extends beyond our noninferiority margin.
Making the raw model wider therefore does not close the gap on BSDS.
Fine-tuning the final DINO block gives no detectable gain over
the frozen bank. SSRB remains stronger on LIDC-hard, while its difference from
this adapted bank on BSDS is unresolved. Paired intervals are in the supplement.

\begin{figure*}[t]
\centering
\includegraphics[width=.95\textwidth]{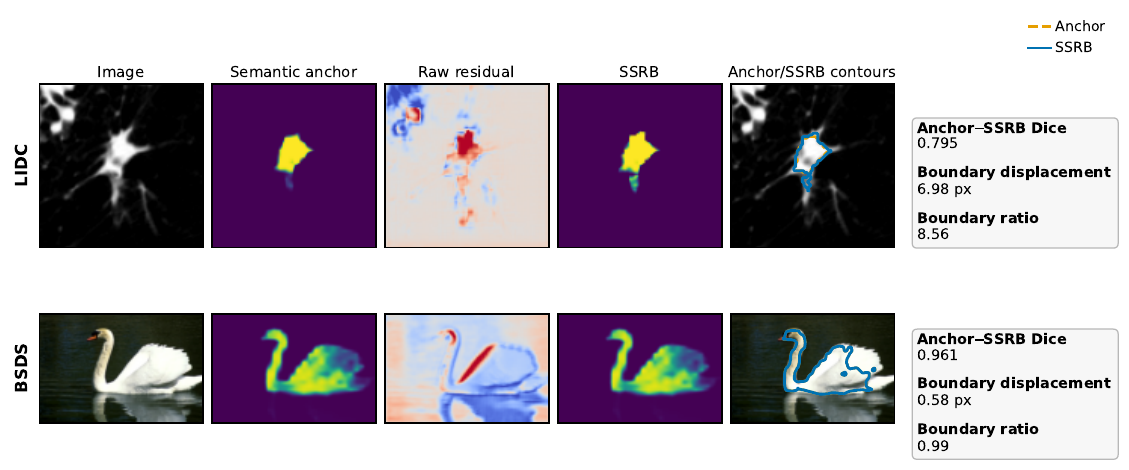}
\caption{\textbf{How the residual changes two representative masks.} On LIDC,
the residual changes the semantic boundary substantially; on BSDS, the swan
example makes preservation of the semantic object mask explicit. The side
values are endpoint means. Boundary ratio is the residual magnitude inside
versus outside the anchor boundary band.}
\label{fig:residual_behavior}
\end{figure*}

Measurements of the outputs agree with Fig.~\ref{fig:residual_behavior}.
Residual logits are small relative to the semantic logits on both datasets.
On LIDC, they concentrate near the semantic boundary, and the final masks
show larger changes in their boundaries than on BSDS. On BSDS, the final
masks remain close to the semantic masks. These observations are consistent
with adjusting nodule extent while retaining the broader regions supplied by
the semantic bank. The supplement reports residual magnitudes, mask overlap,
and boundary displacement.

\paragraph{Backbone and computation.}
Replacing DINOv3 with DINOv2 preserves the main pattern: SSRB improves on its
matching semantic bank on LIDC and LIDC-hard, while BSDS coverage stays close
to that bank. DINOv3 remains the stronger anchor. The supplement reports
paired intervals and measurements of computation. Under the same timing
procedure on an A40, SSRB is slower than either branch alone and than
Probabilistic U-Net because it evaluates both branches for every image.

\begin{figure*}[t]
\centering
\includegraphics[width=.96\textwidth]{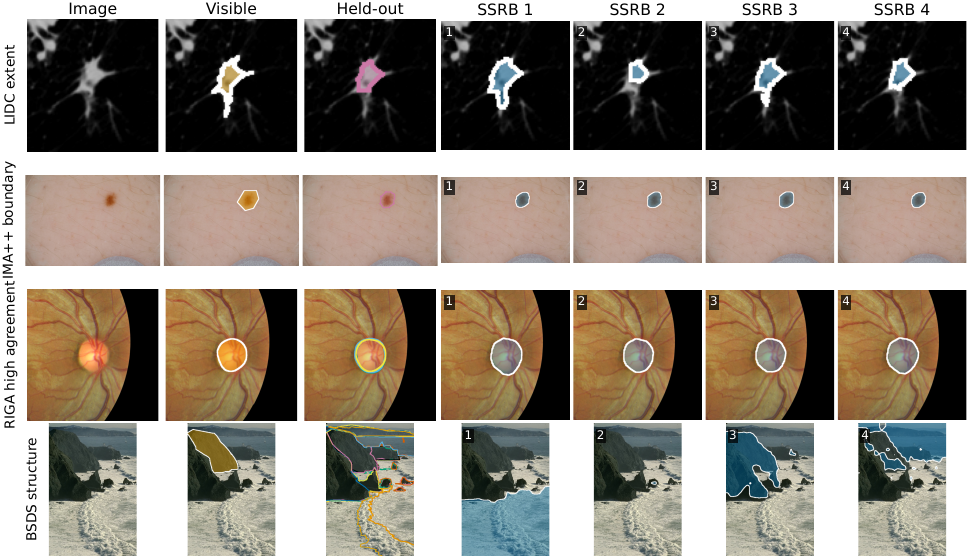}
\caption{\textbf{Qualitative SSRB mask banks.} Each row shows the
image, its fixed visible annotation, held-out contours,
and the first four of eight SSRB masks in score order. Rows were
selected by a prespecified rule based on the median Dice between semantic and SSRB masks. The
examples illustrate local
extent, boundary, high-agreement, and structural variation.}
\label{fig:ssrb_qualitative}
\end{figure*}

Figure~\ref{fig:ssrb_qualitative} shows the ranked mask banks.

\section{Discussion}
\label{sec:discussion}

\paragraph{Complementary representations.}
Raw spatial banks recover variation in nodule extent on LIDC, whereas frozen
semantic banks retain the broader groupings of regions in BSDS. SSRB learns
one spatial residual for each semantic head. On LIDC, small residuals near a
boundary can change the final mask; on BSDS, outputs remain close to their
semantic anchors. The same composition rule supports both behaviors.

\paragraph{Evidence for the construction.}
Learning residuals improves on the tested addition control and gives higher
coverage than the fixed union on BSDS.
Separating the effect of pairing each residual with the same semantic head would require
controls that learn with permuted heads or train all $K$ heads jointly.
Semantic-spatial fusion and residual adaptation provide the broader context
for this construction.

\paragraph{Intended use.}
The bank offers several alternatives that can be retained or inspected.
Its ranking provides an order for inspection, while MLQ@8 measures whether
the bank contains masks close to the held-out annotations. We have not
evaluated a workflow in which users select a mask.

\section{Limitations}
\label{sec:limitations}

The medical endpoints share data: LIDC-hard is a subset of LIDC, and RIGA disc
and cup use the same fundus images. All endpoints were examined during
development, so independent validation on another dataset remains necessary.
BSDS tests alternative groupings of regions, whereas the medical tasks concern
annotations of the same object. LIDC provides crops located from the union
of annotations; extending the method to full images or 3-D volumes requires
further evaluation.

Probabilistic U-Net gives better MLQ@1 on LIDC, LIDC-hard, RIGA-disc, and
RIGA-cup, lower Waste on LIDC, LIDC-hard, and IMA++, and faster inference.
SSRB is also slower than either branch alone. A wider raw model approaches
SSRB on LIDC-hard: the paired interval reaches
-0.0130, beyond our
$-0.010$ noninferiority margin, while remaining substantially below SSRB on
BSDS. Fine-tuning the final DINO block yields no detectable MLQ@8 gain over
the frozen bank; deeper adaptation remains to be studied.

One annotation per image provides limited information about alternatives
that rarely occur across the training population. Recorded disagreement may
also include annotation error~\cite{kimhi2025noisy}. SSRB predicts a finite
set of masks rather than probabilities of annotator choices.

\section{Conclusion}
\label{sec:conclusion}

SSRB learns a spatial correction for each hypothesis in a frozen semantic
bank, using one annotation per image. The semantic bank supplies context for
grouping regions, while the residuals adjust boundaries and extent. Across
the tested medical endpoints and BSDS, the resulting eight masks retain the
complementary strengths of the two representations and improve coverage of
medical annotations over Probabilistic U-Net. Learning these corrections also
improves on the tested addition of independently trained predictions.

The benefit is clearest when several alternatives are retained: the advantage
over Probabilistic U-Net grows as more masks are returned, while the quality of the first mask,
unsupported alternatives, and runtime remain separate considerations. The
results support combining semantic hypotheses with corrections learned
directly from images when training provides limited annotation. Improving
the ordering and reducing unsupported alternatives are natural next steps,
alongside evaluation on independent datasets.

\section*{Acknowledgments}
This work was supported by the Austrian Science Fund (FWF), project DFH 37-N,
Visual Heritage: Visual Analytics and Computer Vision Meet Cultural Heritage.

{
\small
\bibliographystyle{ieeenat_fullname}
\bibliography{main}
}

\end{document}